\documentclass[letterpaper, 10pt, conference]{ieeeconf}
\IEEEoverridecommandlockouts   % needed for \thanks
\usepackage{times}             % Times (Base-14-equivalent, Type 1)
\usepackage{graphicx}
\usepackage{tikz}
\usetikzlibrary{arrows.meta,positioning,fit,backgrounds,calc}
\usepackage{amsmath,amssymb}
\usepackage{booktabs}
\usepackage{cite}
\usepackage{url}
\usepackage{xcolor}
\usepackage{stfloats}
\usepackage{tikz} 
\usetikzlibrary{arrows.meta}

\title{\LARGE \bf
FLINT: Fast Lightweight Inference for Traversability\\
}

\author{William Bonilla$^{1,2,3}$, Maxime Boisvert$^{2}$, David-Alexandre Poissant$^{3}$, David Meger$^{1}$, Louis Petit$^{3}$
\thanks{$^{1}$W. Bonilla and D. Meger are with the School of Computer Science, McGill University, Montreal, QC, Canada. \texttt{\small william.bonillavillatoro@mail.mcgill.ca}}
\thanks{$^{2}$W. Bonilla and M. Boisvert are with Centre de Technologies Avancées (CTA), Sherbrooke, QC, Canada.}
\thanks{$^{3}$W. Bonilla, D.-A. Poissant, and L. Petit are with Université de Sherbrooke, Sherbrooke, QC, Canada.}
}

\begin{document}
\maketitle
\thispagestyle{empty}
\pagestyle{empty}

%=====================================================================
\begin{abstract}

  Navigation in off-road conditions is challenging due to the lack of structure. There is no fixed vocabulary for what is traversable. The traversability depends on both the environment and the embodiment's
  dynamics. Neither of these two variables can be hand-labeled at scale. Thus, traversability has to be learned by the embodiment's own experience. Modern platforms tend to use multiple sensors to estimate traversability and navigate: RGBD cameras, lidar, radar, IMU, with computationally intensive platforms to run inference on neural networks. Against this trend, we propose FLINT, a lightweight traversability estimator: a 21.6M-parameter backbone, 38$\times$ smaller than a comparable foundation-model backbone, that scores higher on held-out terrain probes and runs at 14.7 FPS on CPU alone using a RGB camera has the only sensor. Despite that gap in scale, FLINT produces a cheaper, more accurate costmap than a deployed foundation-model system (WildOS) on 23 of 24 replayed field logs. We compare different self-supervised learning signals and deploy the resulting models on a real platform in closed-loop field trials: the best self-supervised head reaches 99\% autonomy over the route, outperforming a human-label-trained baseline deployed live on the same course. Our results show that heavy sensing and computing are not necessary for traversability estimation.

\end{abstract}


%=====================================================================
\section{Introduction}
Traversability is defined as a robot's ability to cross a terrain region in an admissible state, a concept that depends on the terrain's geometrical and physical characteristics as well as the specific robot's capabilities \cite{gholami2025terrain}. Field deployments in disaster response, agriculture, and planetary exploration all hinge on getting that judgment right, often with no human able to intervene in time~\cite{beycimen2023survey}. Off-road traversability estimation is difficult because of how ambiguous the definition can be \cite{footprints}. It depends on both the terrain and the platform crossing it \cite{wvn, wvn2025}. To reduce this ambiguity, deployed embodiments are equipped with heavy computing and sensing. For example, V-Strong uses only a single camera, but its full pipeline still requires an NVIDIA A100 GPU \cite{vstrong}. WildOS uses a more modest Jetson Orin, but relies on an extensive sensor suite (3 RGBD cameras + a lidar)\cite{wildos}.  
This better sensing alone cannot decide what is traversable and what is not. That capacity requires training labels. The algorithms used require large-scale data to train, and hand-annotated labels at this scale are expensive and slow. The alternative is self-supervised: learn traversability from what the robot has already driven over \cite{wellhausen,badgr,wvn}. 
Self-supervised learning is not trivial. Wellhausen et al. \cite{wellhausen} label from footholds, BADGR \cite{badgr} from future collision, and WVN \cite{wvn}from online visual similarity. Each is a different definition of the same label: a foothold marks a single point the robot's foot safely touched, a collision label marks a stretch the robot drove without crashing, and a visual-similarity label marks anything that looks like ground already crossed. None of these definitions is obviously the right one, and the choice shapes what the model actually learns. The generated labels do not just describe the terrain, it inherits the capability of the platform that collected them. A robot that can crush a brush will label the brush as traversable, and that label transfers unchanged to a smaller robot that cannot. 

This paper makes two contributions:
\begin{itemize}
    \item \textbf{A real-time, CPU-only, single-camera traversability estimator.} A frozen backbone and a lightweight trained head produce dense traversability costmaps without a GPU, lidar, or depth sensor, at a smaller sensing and compute footprint than comparable field systems.
    \item \textbf{A controlled comparison of self-supervision methods.}
    We compare 3 ways of turning a driven corridor into training
    labels, on identical data and backbone, against a human-labeled
    baseline, and evaluate them zero-shot, in closed-loop replay, and
    live in the field.
\end{itemize}
We validate both contributions with real robot experiments on a Clearpath Husky A200, using only a single RGB camera, in mixed forest and gravel terrain. In a controlled outdoor deployment, holding route, day, and planner fixed and varying only the traversability head, our best self-supervised heads require fewer interventions than a human-labeled baseline trained specifically for this task. This advantage is not an artifact of one label source: propagating the driven corridor to the rest of the coherent traversable surface, rather than labeling only the corridor itself, is what separates the strongest self-supervised heads from the baseline, and the same ranking reproduces on an independent, offline zero-shot benchmark — pointing to the propagation mechanism itself, rather than any particular label generator, as the source of the gain. Finally, replayed offline against Husky field logs, our heads produce a cheaper driven-path costmap than a deployed foundation-model system, at a fraction of its latency and memory, all from a pipeline that runs on CPU alone showing the result holds even against much larger backbones, not just a human baseline.

\section{Related Work}
\label{sec:related}

Numerous strategies have emerged for teaching robots what terrain they can cross, many built on a different notion of self-supervision, sensor suite, and backbone. We group this body of work around two questions: what signal is used to generate labels from a robot's own experience, and how much sensing and compute that signal demands to run in the field. Finally, we cover the pre-trained backbones and datasets that anchor these methods.

\subsection{Self-supervised labels for traversability}
Self-supervision avoids hand-labeling by treating the terrain a platform has already crossed as a positive traversabillity label, and numerous approaches have been proposed for off-road autonomy \cite{wellhausen, badgr, wayfast, wayfaster,footprints, castro2023howdoesitfeel, zurn2020acoustic, wvn, wvn2025, uavterrain, vstrong}. These efforts employ a wide range of modalities to generate that signal. Wellhausen et al.~\cite{wellhausen} derive it from contact-reprojected footholds a legged robot's own feet already crossed safely; BADGR~\cite{badgr} instead supervises from the opposite polarity, learning to predict future collision and bumpiness events from IMU and lidar so a policy can steer away from the terrain that produced them, rather than toward terrain already proven safe. WayFAST~\cite{wayfast}, WayFASTER~\cite{wayfaster}, and Jeon et al.~\cite{footprints} instead track a traction or footprint estimate over RGB-D. Castro et al.~\cite{castro2023howdoesitfeel} and Zürn et al.~\cite{zurn2020acoustic} skip vision entirely, using an IMU-conditioned cost and acoustic terrain sound; Fortin et al.~\cite{uavterrain} fuse vibration and energy consumption from a UAV-aligned proprioceptive trace. Two are methodologically closest to the strategies we compare later: Wild Visual Navigation (WVN)~\cite{wvn,wvn2025} propagates driven labels to visually similar regions from a short teleoperated demonstration, and V-STRONG~\cite{vstrong} derives positive and negative samples from SAM-segmented instances and the driven trajectory. Even the same underlying signal a camera watching where the robot drove yields different information depending on how it is processed: WVN treats it as a similarity cue between regions, V-STRONG as a set of discrete positive and negative instances. Despite this diversity, none have been compared directly on identical data, backbone, and protocol.

\subsection{Computing and sensing}
  Off-road systems increasingly trade sensing and compute for robustness. V-STRONG~\cite{vstrong} and Wild Visual Navigation~\cite{wvn,wvn2025} both need only a single camera, but neither runs without a GPU: V-STRONG's full pipeline reaches 6~Hz on a single NVIDIA A100, and WVN likewise requires a GPU to run its DINO backbone in real time. WayFAST~\cite{wayfast} and WayFASTER~\cite{wayfaster} target the same real-time, single-robot setting we do, running on an embedded Jetson GPU rather than a workstation-class accelerator, but still depend on a depth channel: both track their traction/footprint estimate over RGB-D, whereas our system runs on a single RGB stream and no GPU at all. WildOS~\cite{wildos} adds heavy sensing on top of heavy compute: an Ouster OS0-128 lidar, a VectorNav VN-100 IMU, and three Intel RealSense D455 RGB-D cameras feed a foundation-model backbone on a Jetson AGX Orin, yet its own reported inference rate is only 1.4~Hz on that hardware. Boxi~\cite{Frey-Tuna-Fu-RSS-25} goes further still: two lidars, ten RGB cameras, an RGB-D camera, seven IMUs, and a dual-antenna RTK-GNSS receiver twenty-one sensors on one payload.

% \subsection{Cross-Platform capability transfer}
% Most self-supervised traversability systems avoid the cross-platform question by construction: the robot that generates the label is the robot that drives on it. Wild Visual Navigation~\cite{wvn,wvn2025} adapts online from a short teleoperated demonstration on the deploying platform itself, so transfer to a different platform is never tested. V-STRONG~\cite{vstrong} instead pre-trains once on large-scale data and generalizes zero-shot, exactly the setting where platform inheritance stays invisible until it causes a collision, since no benchmark pixel-metric certifies against it. SALON~\cite{salon} is the closest published attempt to close this gap directly: a frozen DINOv2 backbone, proprioception, and a single human-clicked hard-negative seed let it adapt in seconds across platforms as different as a wheelchair, a quadruped, and an ATV, evidence that a small amount of target-platform data closes the cross-platform gap. None of these measure how far the gap extends without any target-platform data.
  
\subsection{Datasets and backbones}
Frozen vision foundation models increasingly serve as backbones for downstream perception without any task-specific fine-tuning. DINOv3~\cite{dinov3} in particular produces dense, patch-level features that transfer across domains without further training, making it attractive as a shared encoder for comparing different downstream strategies on equal footing. Segmentation tools built on this kind of backbone offer complementary views of the same scene: SAM3~\cite{sam3} provides class-agnostic instance masks with strong object-level boundaries, while STEGO~\cite{stego} instead produces unsupervised semantic clusters, grouping pixels by learned visual similarity rather than discrete object instances. Together, these tools make it possible to generate structured spatial priors instances or semantic groupings without any manual annotation or model fine-tuning. On the data side, TartanDrive 2.0~\cite{tartandrive2} offers large-scale, real-world ATV driving data well suited to self-supervised label generation, while RELLIS-3D~\cite{rellis3d} offers densely human-annotated UGV terrain imagery, making it a natural benchmark for evaluating generalization across label sources.

\section{Method}
In this section, we describe how the same driven corridor is turned into training labels under 3 propagation strategies, and how each is used to supervise a lightweight head on a shared frozen backbone. We first detail trajectory reprojection, followed by SAM3 instance propagation and STEGO cluster propagation, each defining positive and negative regions differently from the same underlying signal. We then describe how these labels are used to train the traversability head and how its output is converted into a costmap for deployment.

\subsection{Self-Supervised Label Generation} 
\begin{figure*}[!t]
  \centering
  \includegraphics[width=\textwidth]{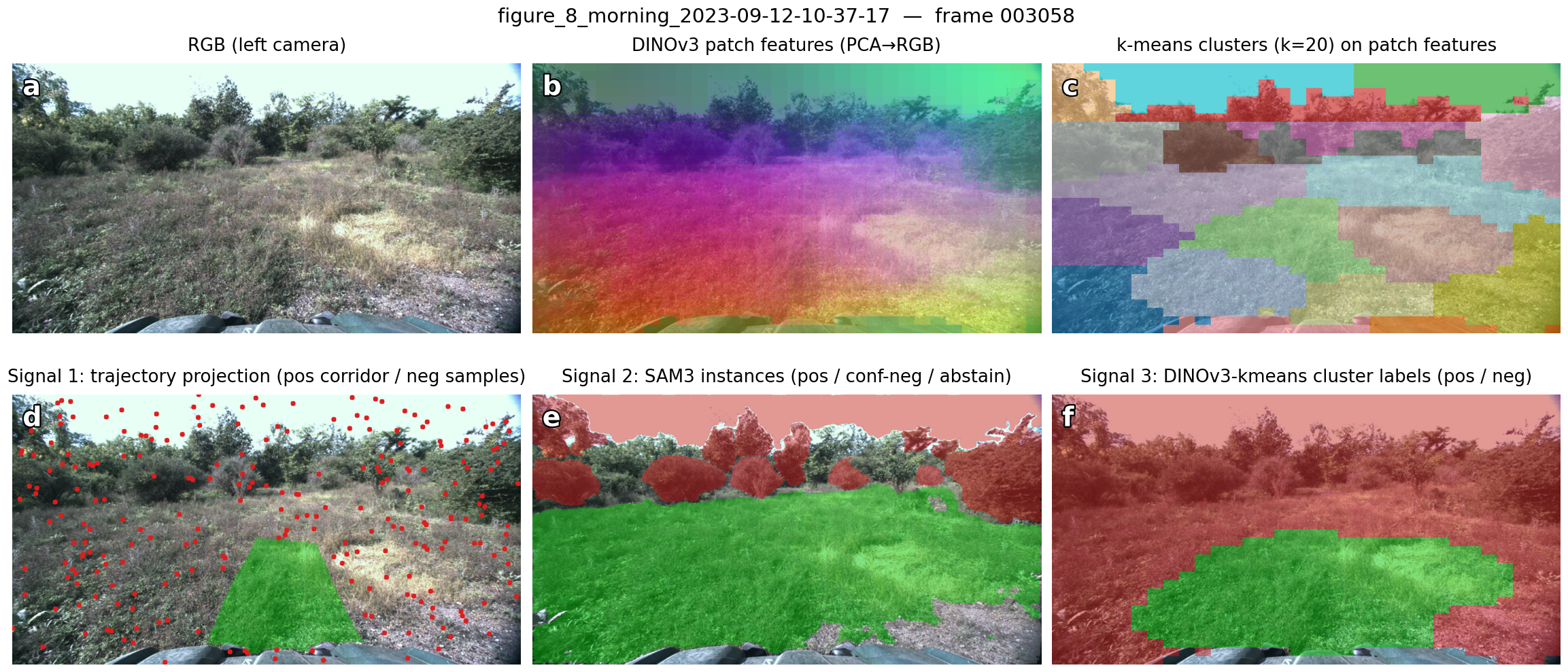}
\caption{Label generation on one TartanDrive~2.0 frame (open grass
  field, \texttt{figure\_8\_morning} run, frame 003058). (a) RGB input;
  (b) DINOv3 patch features (PCA$\to$RGB); (c) k-means clusters (k=20)
  on those features. The three self-supervision signals: (d)
  \textit{traj} — reprojected 2\,s wheel corridor (green) with sampled
  negatives (red points) outside a dilated margin; (e) \textit{sam3} —
  SAM3 instances classified by corridor overlap (green positive, red
  confident-negative sky/tree, unshaded abstain), which propagates the
  narrow corridor to the entire drivable surface; (f) \textit{stego} —
  clusters labeled by corridor overlap on the patch grid. The same drive
  supervises each head very differently: the corridor band alone (d),
  the full coherent surface (e), or feature-space clusters (f).}
  \label{fig:signals}
\end{figure*}

 In this work, we use three propagation strategies: trajectory reprojection (\textit{traj}), SAM3 instance propagation (\textit{sam3}), and STEGO cluster propagation (\textit{stego}),  illustrated in Fig.~\ref{fig:signals}
on a single driven frame. Each of the three propagation strategies turns the same driven corridor into a training label for its own version of the MLP head on the frozen DINOv3~\cite{dinov3} backbone. Labels are generated from TartanDrive~2.0~\cite{tartandrive2}, using each run's own super-odometry and camera calibration to reproject the vehicle's path. \textbf{Trajectory reprojection}, in the spirit of Wild Visual Navigation's footprint-based supervision~\cite{wvn,wvn2025}, projects the vehicle's future 2.0\,s wheel corridor into the image as the positive region; negatives are sampled outside a dilated margin around that corridor, excluding sky and an automatically fitted vehicle-hood mask, and are binned directly onto the 32$\times$32 patch grid.\textbf{SAM3 instance propagation}, following V-STRONG's use of segmentation to extend a sparse trajectory signal~\cite{vstrong}, uses SAM3's~\cite{sam3} text-prompted instance segmentation over an 11-term terrain vocabulary (dirt trail, rocky ground, smooth dirt road, grass, tall grass, bush, tree, mud, puddle, rock, sky); any instance whose area overlaps the driven corridor by at least 5\% is labeled positive in full, propagating the sparse trajectory signal to
the entire coherent surface it lies on. Only \texttt{sky} is confident negative, labeled negative unconditionally regardless of overlap; every other term that falls short of the overlap threshold abstains rather than being labeled negative, since not having been driven on during one pass is not evidence that, say, a patch of grass is non-traversable in general.

\subsection{Head Training}

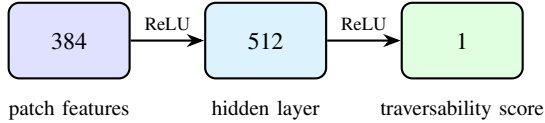
\begin{figure}[t]
\centering
\begin{tikzpicture}[
    ->, >=Stealth, thick,
    box/.style={rectangle, draw=black, rounded corners, minimum width=1.6cm, minimum height=1cm, font=\small},
    input box/.style={box, fill=blue!12},
    hidden box/.style={box, fill=cyan!12},
    output box/.style={box, fill=green!12},
    lbl/.style={font=\footnotesize, text centered}
]

    \node[input box] (in) at (0, 0) {384};
    \node[hidden box] (h) at (2.6, 0) {512};
    \node[output box] (out) at (5.2, 0) {1};

    \draw (in) -- (h) node[midway, above, font=\scriptsize] {ReLU};
    \draw (h) -- (out) node[midway, above, font=\scriptsize] {ReLU};

    \node[lbl, below=0.15cm of in] {patch features};
    \node[lbl, below=0.15cm of h] {hidden layer};
    \node[lbl, below=0.15cm of out] {traversability score};

\end{tikzpicture}
\caption{MLP head architecture: a 384-d frozen DINOv3 patch feature is mapped through a 512-unit hidden layer to a scalar traversability score.}
\label{fig:mlp_head}
\end{figure}

  \begin{enumerate}
      \item \textbf{Data and split.}The training data is split by run rather than by frame: adjacent frames from the same drive are highly correlated, so a frame-level split would let a head see near-duplicates of its own validation data during training and report inflated performance. Every head instead trains on the same 49 TartanDrive~2.0 runs (104,083 valid frames of 119,020 raw, stationary segments excluded), split 85/15 \textit{by run} with a fixed seed, so no run leaks between train and validation.

      \item \textbf{Optimization.} Each head is the same MLP head (Fig.~\ref{fig:mlp_head}) with 197,633 trainable parameters, 0.9\% of the frozen backbone trained with BCEWithLogitsLoss and AdamW (learning rate $10^{-4}$, weight decay 0.01, cosine-annealed) for 20 epochs on a single 12\,GB desktop GPU (RTX 5070).

      \item \textbf{Per-head cost.} \textit{sam3} and \textit{stego} train MLP-only on pre-extracted per-frame features (batch 256, minutes per run); \textit{traj} recomputes the frozen backbone at every step, since its labels are purely geometric and not cached.
\end{enumerate} 

\subsection{System Overview}
\begin{figure*}[!t]
\centering
\includegraphics[width=\textwidth]{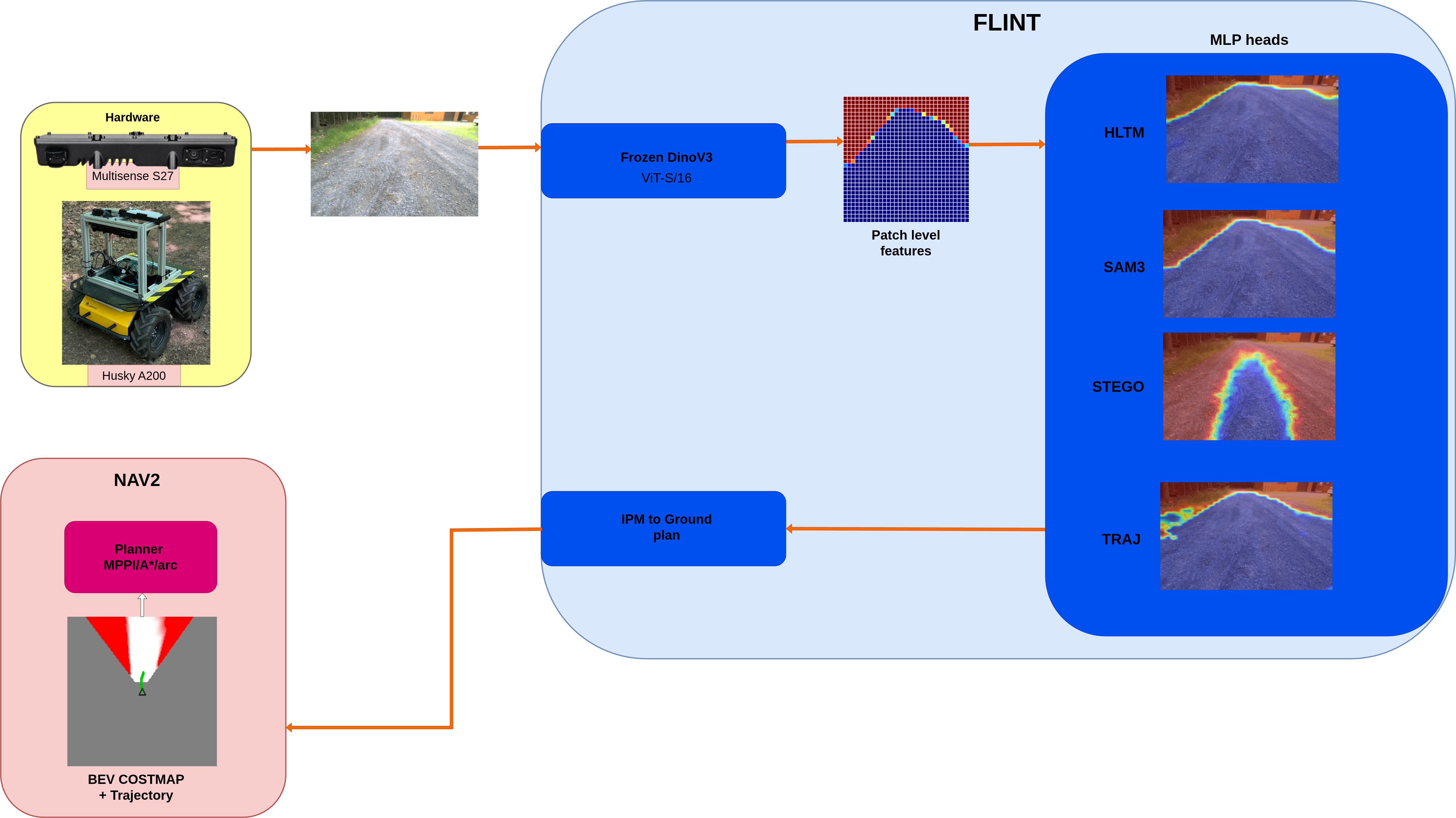}
\caption{\textbf{System Overview.} A single RGB frame captured by the Husky A200's Multisense S27 stereo camera is processed by a frozen DINOv3 (ViT-S/16) backbone into patch-level features. Four lightweight MLP heads (HLTM, SAM3, STEGO, TRAJ) independently propagate these features into a dense traversability probability map, shown here for each head on real field data (blue = high traversability probability, orange = low). The active head's probabilities are passed through an inverse perspective mapping (IPM) module to produce a ground-plane occupancy grid, which Nav2 consumes as a BEV costmap to generate a planned trajectory (white corridor) via its MPPI/A*/arc planner. Colored regions denote organizational groupings only (hardware, FLINT, Nav2), not a claim of novelty.}
\label{fig:system-overview}
\end{figure*}

FLINT itself comprises the frozen backbone, the active propagation
  strategy's MLP head, and the IPM projection step; Nav2 and the
  underlying hardware are not part of our contribution, and are included
  in Fig.~\ref{fig:system-overview} only to show the complete deployed
  system.

  \begin{itemize}
      \item \textbf{Hardware.} The robot is equipped with a MultiSense S27 stereo camera, already mounted on the Husky prior to this work rather than selected for it, and standard wheel encoders for odometry. We use only the camera's single rectified RGB stream and never its depth output.

      \item \textbf{Costmap.} Each frame is processed by a frozen DINOv3 ViT-S/16 backbone into a 32$\times$32 grid of patch-level features, from which the active propagation strategy's head predicts a per-patch traversability probability. These probabilities are projected onto the ground plane via inverse perspective mapping and handed to Nav2~\cite{macenski2020marathon2,macenski2023survey}  as an OccupancyGrid; Nav2 owns costmap construction, planning, and low-level control from that point on.

      \item \textbf{Planner.} The costmap is planner-agnostic; any Nav2-compatible planner can consume it without retraining or modifying FLINT. 

      \item \textbf{Onboard Computation.} In deployment, exactly one head is active at a time. The entire forward pass, backbone and that one head, runs on CPU alone at 68\,ms/frame (14.7 FPS). Because the backbone is frozen and shared across heads, its 68\,ms cost is paid once regardless of how many traversability heads are evaluated on the same frame; each additional lightweight head adds negligible overhead on top, a property this paper exploits only for the offline side-by-side comparisons of Sec.~\ref{sec:baselines} \ref{sec:head-to-head-self}, not in the live field system, which always runs a single head.

  \end{itemize}

 \subsection{Human-Labeled Baseline}
  We include \texttt{HLTM} to establish how far self-supervised labels
  can close the gap to human annotation, the offline accuracy ceiling
  this paper's self-supervised heads are measured against;
  Sec.~\ref{sec:heads-vs-hltm} analyzes this comparison directly, both
  in live closed-loop deployment and offline zero-shot evaluation.
  \texttt{HLTM} shares the same frozen DINOv3 backbone and MLP head as
  the three self-supervised heads above, but its label comes from human
  annotation rather than driven-corridor propagation: dense per-pixel
  traversability masks pooled across three human-labeled datasets
  (Yamaha~\cite{maturana2018yamaha}, 1,076 images;
  CAVS~\cite{dabbiru2025cassed}, 677 images; Great
  Outdoors~\cite{go2025dataset}, 1,199 images), collapsed to a binary
  traversable/non-traversable label under the same convention used to
  build the RELLIS-3D evaluation ground truth. RELLIS-3D itself is held
  out entirely from this training pool, so \texttt{HLTM}'s RELLIS-3D
  evaluation is genuinely zero-shot, exactly like the other three heads.

\section{Experiments}
  This section evaluates FLINT along three axes: how it compares against
  heavier foundation-model baselines, how its self-supervised heads
  compare against a human-labeled baseline in live deployment, and why
  the winning self-supervised heads win. Sec.~\ref{sec:baselines} first
  establishes that FLINT is competitive with a much larger
  foundation-model system, before turning to comparisons among our own
  heads. Sec.~\ref{sec:heads-vs-hltm} then presents this paper's
  headline field result: two self-supervised heads outperform a
  human-labeled baseline in live closed-loop deployment.
  Sec.~\ref{sec:head-to-head-self} explains that result mechanistically,
  bringing the weakest self-supervised head back into the comparison to
  show what separates it from the two winners. Sec.~\ref{sec:cpu-bench}
  closes with the compute cost behind all of the above, confirming the
  entire pipeline sustained real-time inference on CPU alone throughout
  every field deployment.

\subsection{Offline Comparison Against Foundation-Model Baselines}
  \label{sec:baselines}
  Before comparing self-supervision strategies against each other, we first ask whether FLINT is competitive with state-of-the-art traversability estimators at all. Live, closed-loop deployment of two natural baselines was not feasible on our platform: WildOS~\cite{wildos}'s full stack assumes a sensor suite (3 RGB-D cameras + a lidar) our Husky A200 does not carry, and Wild Visual Navigation (WVN)~\cite{wvn,wvn2025}'s learning node is implemented in ROS1, with no straightforward path onto our ROS2/Nav2 stack. Instead, we ran WildOS's own inference node for traversability estimation (ExploRFM) directly against RGB alone, excluding its lidar and elevation-mapping stages, to isolate the vision module both systems actually share, and built a native, ROS-independent reimplementation of WVN's core online-learning algorithm, run in shadow mode (no actuation) but still taking real gradient steps from the log's own velocity-tracking error. Both were replayed against 24 Husky field logs spanning five collection days, alongside our own three heads served from one shared-backbone process, and every resulting costmap was scored with the same threshold-free driven-cost percentile used throughout this paper: the actually-driven path is a fact independent of any head's own calibration, so the metric holds even though WildOS, WVN, and our three heads disagree about what to do with unfamiliar terrain.

  \begin{table}[!t]
  \centering
  \caption{Offline comparison against foundation-model baselines}
  \label{tab:baselines}
  \resizebox{\columnwidth}{!}{%
  \begin{tabular}{lcccc}
  \toprule
  System & Mean pct.\ & $<$WildOS & ms/frame & GPU MB \\
  \midrule
  stego                          & \textbf{18.1} & 23/24 & 70.6  & 391 \\
  traj                           & 23.8 & 23/24 & 70.6  & 391 \\
  sam3                           & 26.5 & 22/24 & 70.6  & 391 \\
  \midrule
  WildOS~\cite{wildos}           & 41.0 & ---   & 405.3 & 1045 \\
  WVN (shadow)~\cite{wvn,wvn2025} & 54.2 & 4/24  & 44.0  & 382 \\
  \bottomrule
  \end{tabular}%
  }
  \end{table}

  Table~\ref{tab:baselines} gives the result: all three of our heads
  produce a cheaper driven-path costmap than WildOS in the large majority
  of logs (\texttt{traj}/\texttt{stego} 23/24, \texttt{sam3} 22/24), at
  5.7$\times$ lower latency on the deployed Jetson Xavier and roughly a third of
  its GPU memory sharing one 70.6\,ms backbone pass across all three
  heads where WildOS pays 405.3\,ms for one. WVN-shadow is competitive on
  latency and memory but beats WildOS on only 4 of 24 logs, sitting close
  to the random-path level on several of them, a caveat worth noting
  before treating it as a strong baseline elsewhere. This establishes that FLINT is not merely cheap: on this deployed-costmap metric it is also competitive with, and typically better than, a foundation-model system built for the same task,
  despite FLINT's backbone being 38$\times$ smaller and needing a
  lighter sensor suite. With that established, the rest of this
  section turns to the comparison this paper is actually about: which
  way of turning a driven corridor into a self-supervised label works
  best.

\subsection{Self-Supervised Heads vs.\ Human-Labeled Baseline}
  \label{sec:heads-vs-hltm}

  Before presenting this section's result, one definition needs
  to be fixed: what counts as an intervention. We define it as any
  interval in which the operator disengages autonomous mode via the
  planner's deadman switch, regardless of cause. We further distinguish two categories by manually reviewing the
  costmap and driven context at each episode, a \textbf{minor}
  intervention being a brief correction issued while the active head's
  traversability estimate was locally correct, needed only because the
  planner could not resolve a sharp turn or trail junction from within
  the costmap's current field of view, and a \textbf{major} intervention
  being a takeover triggered by an actual traversability
  misclassification, where the active head's costmap marked an unsafe
  region as traversable (or the reverse), the planner routed the robot
  toward the resulting hazard, and the operator had to redirect it away,
  a distinction made manually from the recorded costmap and video rather
  than an automatically logged signal, and one that reported
  intervention counts and intervention-free distances below aggregate
  across unless noted otherwise
  
\begin{figure*}[t]
  \centering
  \includegraphics[width=\textwidth]{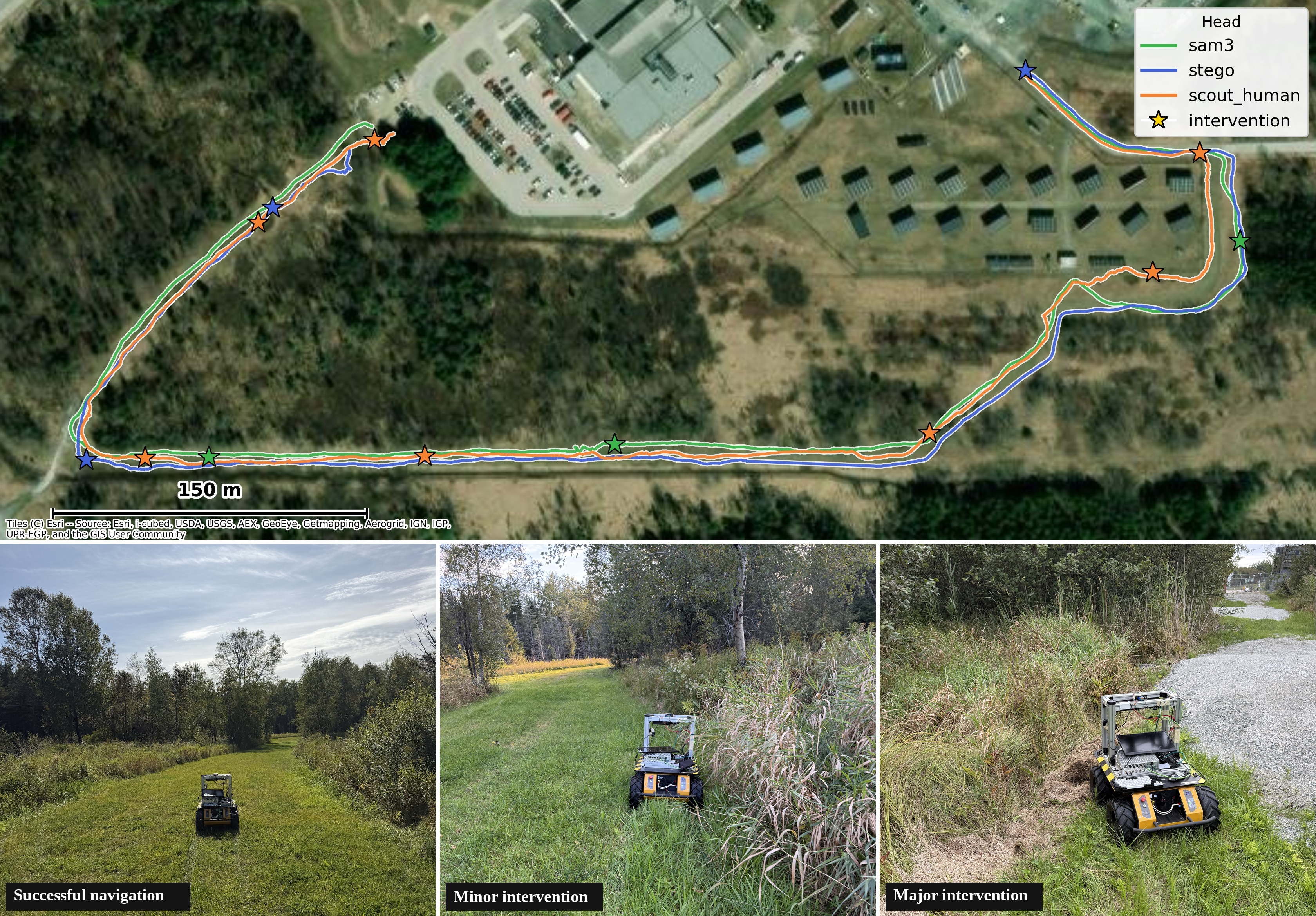}
  \caption{GPS trajectories for stego, sam3 and HLTM heads on the
  identical route and planner, overlaid on a shared basemap. Clustered
  operator-intervention episodes are marked per head. Added below are
  example interventions: successful navigation, a minor intervention
  near the trail edge, and a major intervention into high grass.}
  \label{fig:multihead-traj}
  \end{figure*}

  To isolate the effect of the traversability head from all other
  system variables, we deployed all four candidate heads (\texttt{traj},
  \texttt{sam3}, \texttt{stego}, and a human-label-trained model,
  \texttt{HLTM}) on the same route, on the same day, using the identical
  A* planner and costmap interface; only the head producing the
  traversability estimate differed between runs. Across the four runs,
  the robot covered a combined 4.47\,km, accumulating 119.1\,min of
  continuous autonomous-mode CPU-only inference with no dropped frames.
  Fig.~\ref{fig:multihead-traj} overlays \texttt{sam3}, \texttt{stego},
  and \texttt{HLTM}'s driven GPS trajectories on a shared basemap, with
  clustered operator-intervention episodes marked; \texttt{traj}'s
  trajectory is omitted from this figure for legibility (frequent
  interventions fragmented its track into many short, overlapping
  segments) but its full results are in Table~\ref{tab:heads-vs-hltm}
  along with the other three.

 Table~\ref{tab:heads-vs-hltm} gives the result of this paper's field
  comparison: the two self-supervised heads, \texttt{sam3} and
  \texttt{stego}, both outperformed \texttt{HLTM}. \texttt{sam3}
  required the fewest interventions relative to distance and reached the
  highest autonomy rate (99.0\%), while \texttt{stego} produced the
  single longest intervention-free stretch of any head, 809.9\,m
  (21.9\,min) continuous. Notably, both self-supervised heads
  outperformed \texttt{HLTM} in live closed-loop deployment (96.3\%
  autonomy, 7 interventions), despite \texttt{HLTM} being the strongest
  head, zero-shot, on the offline RELLIS-3D benchmark
  (Table~\ref{tab:rellis}). We evaluate on RELLIS-3D~\cite{rellis3d} both because V-STRONG~\cite{vstrong} reports its own results on the same dataset, letting us situate our numbers against a prior system on common ground, and because its
  dense, human-annotated ground truth makes it a default standard
  benchmark for off-road traversability. Self-supervision did not just approach
  human-labeled quality; it exceeded it once deployed on the platform
  that generated its own labels. This ranking holds up under a second,
  independent measure. We define driven-cost percentile as
  \begin{equation}
  p = \frac{\left|\{\, i : c_i \le c_{\text{driven}} \,\}\right|}{N} \times 100,
  \label{eq:driven-cost-percentile}
  \end{equation}
  where $c_{\text{driven}}$ is the cost of the head's actually-driven
  path and $\{c_1,\dots,c_N\}$ ($N{=}100$) are the costs of random,
  kinematically-feasible alternative paths sampled from the same start
  over the same horizon; a value of 0 means the driven path was cheaper
  than every alternative, 50 is indistinguishable from a random path,
  and 100 means it was the most expensive path tried. On the same
  deployment, this places \texttt{sam3} (13.9\%) ahead of
  \texttt{scout\_human/HLTM} (15.0\%), itself ahead of \texttt{stego}
  (16.6\%); this is the same best-to-worst ordering, among these three
  heads, as the offline zero-shot RELLIS-3D evaluation. Two
  independently-designed protocols, one offline and one live
  closed-loop, agreeing on the same ranking is corroborating evidence
  rather than coincidence.

  \begin{table}[!t]
  \caption{Zero-shot RELLIS-3D evaluation (IoU at fixed threshold $\tau=0.5$).}
  \label{tab:rellis}
  \centering
  \resizebox{\columnwidth}{!}{%
  \begin{tabular}{llccc}
  \toprule
  model & superv. & AUROC & AUPRC & IoU \\
  \midrule
  traj  & self, geom.  & 0.639 & 0.447 & 0.366  \\
  stego & self, clust. & 0.948 & 0.911 & 0.418  \\
  sam3  & self, SAM3   & \textbf{0.956} & \textbf{0.918} & \textbf{0.749}  \\
  \midrule
  HLTM  & human         & 0.985 & 0.963 & 0.824  \\
  \bottomrule
  \end{tabular}%
  }
  \end{table}

  \begin{table}[!t]
  \centering
  \caption{Self-supervised heads vs.\ the human-labeled baseline,
  identical route and planner.}
  \label{tab:heads-vs-hltm}
  \resizebox{\columnwidth}{!}{%
  \begin{tabular}{lccccc}
  \toprule
  Head & Dist. (m) & Auton. \% & Longest (m) & Longest (s) & Interv. \\
  \midrule
  sam3   & 1125.7 & \textbf{99.0} & 398.3 & 639.8  & \textbf{3} \\
  stego  & 1059.7 & 97.7 & \textbf{809.9} & \textbf{1313.4} & \textbf{3} \\
  HLTM   & 1114.5 & 96.3 & 279.8 & 454.8  & 7 \\
  traj   & 1165.2 & 87.5 & 197.0 & 315.8  & 8 \\
  \bottomrule
  \end{tabular}%
  }
  \end{table}
\subsection{Head-to-Head Among Self-Supervised Signals}
\label{sec:head-to-head-self}

  \texttt{traj} was left out of Fig.~\ref{fig:multihead-traj} so that
  result would land cleanly. \texttt{traj} was the weakest head by every
  measure: 87.5\% autonomy, 8 interventions live, and 22.1\% on the
  driven-cost-percentile analysis above, worst on both protocols, and
  worst in the offline zero-shot RELLIS-3D ranking. The reason is what
  \texttt{traj} does \emph{not} do: its label only ever covers the
  driven corridor itself, a fixed-width band the vehicle happened to
  occupy, and never propagates to the rest of the visibly coherent
  traversable surface around it. \texttt{sam3} and \texttt{stego} are
  best understood as a more structured version of that same raw signal:
  \texttt{sam3} propagates it through SAM3's instance segmentation
  masks, and \texttt{stego} through STEGO's superpixel-like feature
  clusters, and it is that additional structure, not a better label
  source, that lets them close the gap to \texttt{HLTM} and then exceed
  it. Put differently, Sec.~\ref{sec:heads-vs-hltm}'s headline result is
  not a property of self-supervision generically; it is a property
  specifically of self-supervision that propagates beyond the driven
  band, and \texttt{traj} is the control condition that shows what is
  lost without it. This is also evidence for a broader point:
  traversability is a property of a region, not just the exact footprint
  a vehicle happened to occupy, and \texttt{traj}'s failure is the
  direct demonstration of that. That \texttt{sam3} and \texttt{stego}
  can recover this broader region from nothing but a frozen backbone and
  a driven corridor is itself consistent with a known property of
  DINO-style self-supervised features: they carry explicit segmentation
  structure that does not emerge as clearly from supervised training or
  direct segmentation objectives~\cite{caron2021emerging}, and our
  result is one more data point for that claim in a domain, off-road
  terrain, far from where it was first observed.

\begin{figure}[t]
  \centering
  \includegraphics[width=\linewidth]{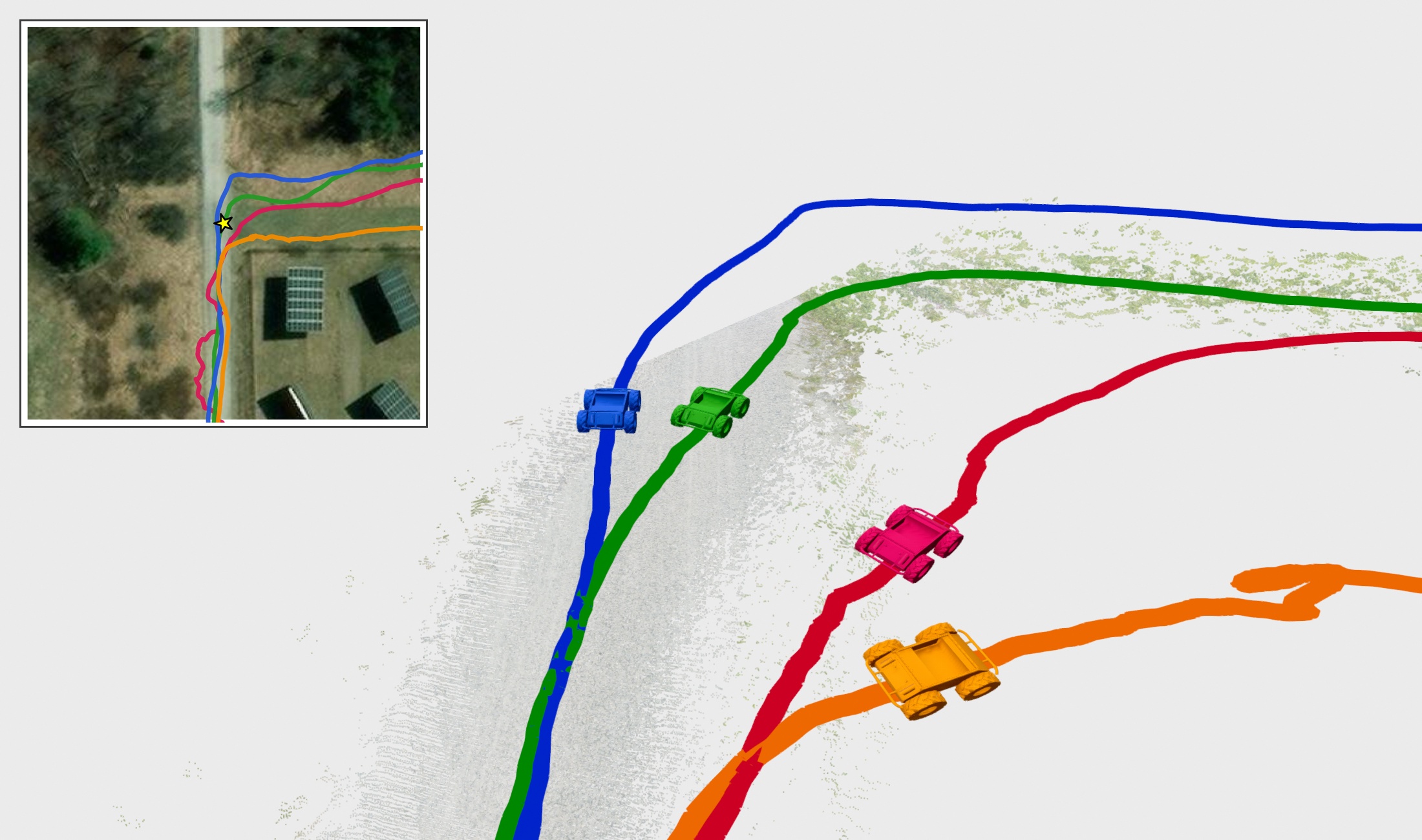}
  \caption{3D reconstruction of the route's sharpest turn, colored by head, with satellite insent showing the driven path}
  \label{fig:rtabmap-safespot}
  \end{figure}

 Fig.~\ref{fig:rtabmap-safespot} makes a related point concretely rather
  than statistically. All four heads find this turn traversable,
  \texttt{stego}, \texttt{sam3}, and \texttt{traj} cross it without
  intervention, and \texttt{HLTM} needs only one small correction, yet
  each settles onto a physically different line through the identical
  turn, from \texttt{stego}'s tight inside cut to \texttt{HLTM}'s wide
  outside arc. There is no single correct line here: all four solutions
  work. What differs is not whether the terrain is traversable but how
  each head's training shapes the specific path chosen through it,
  evidence that the training signal leaves a real, physical fingerprint
  on behavior even when every resulting trajectory succeeds.

\subsection{CPU-Only Compute Benchmark}
 \label{sec:cpu-bench}
  Unlike a bench-only latency figure (68.2\,ms/frame, 14.7\,FPS), this
  throughput was sustained live: FLINT ran CPU-only inference in the
  robot's real-time control loop for the entirety of all four field
  deployments, 119.1\,min of continuous autonomous-mode inference over
  4.47\,km combined, with no dropped frames or fallback to a lower rate.
  The same inference code, run on both devices of one desktop machine,
  confirms this is not a lower bound in principle: 68.2\,ms/frame
  (14.7\,FPS) on CPU alone (16 threads) versus 12.3\,ms/frame
  (81.5\,FPS) on its GPU (RTX 5070), comfortably above the field
  system's real-time floor either way. GPU-free deployment is a live
  option on the real-time control computer, not just something that
  happens to work on desktop hardware.

  We separately benchmarked WVN's~\cite{wvn,wvn2025} own
  feature-extraction pipeline on the \texttt{sam3} field bags, forced
  onto CPU rather than its default GPU path: it processed frames at a
  mean 832.7\,ms/frame (1.20\,Hz), \textbf{12.2$\times$ slower than
  FLINT's 68.2\,ms/frame} on the same CPU hardware, a lower bound on its
  true cost since its online-learning step was not confirmed active
  during measurement.
\section{Conclusion}
We presented FLINT, a traversability estimator built on a frozen,
21.6M-parameter backbone 38$\times$ smaller than a comparable
foundation-model backbone that runs at 14.7\,FPS on CPU alone,
requiring no GPU for inference, and roughly 12$\times$ faster than a
comparable prior system's feature-extraction pipeline when both are
run CPU-only. The system sustained real-time CPU-only inference live, in the robot's control
loop, for 119.1\,min of continuous autonomous driving across 4.47\,km.
We compared three self-supervised label sources against a
human-labeled baseline in a controlled head-to-head field deployment,
holding the route and planner fixed and varying only the
traversability head. The best self-supervised head reached 99.0\% autonomy and outperformed the human-labeled baseline in live deployment, despite the latter beating every self-supervised head zero-shot on RELLIS-3D. An independent driven-cost-percentile analysis of the same deployment agrees with our offline zero-shot ranking on RELLIS-3D,
corroborating the result across two independently-designed
evaluations. A qualitative comparison at a single shared,
zero-intervention waypoint further shows that heads can agree on safe
outcomes while disagreeing on the underlying traversability estimate; 
multiple valid representations can each support safe autonomy
without converging to the same judgment. These results indicate that
reliable traversability estimation does not require heavy sensing or
compute, and that self-supervised labels collected by the deploying platform itself can match or exceed hand-labeled supervision once evaluated in closed loop. Limitations of this study include a single test route and single-day conditions. Future work includes testing across a wider range of terrain and conditions, and closing the loop with online adaptation of the traversability head during deployment.

\section*{Acknowledgment}
Portions of this manuscript's prose (revision throughout all sections), figure-generation, and table formatting were produced with the assistance of Claude (Anthropic), under continuous author direction and review. All experimental design, data collection, code, results, and technical claims were carried out and verified by the authors.

\bibliographystyle{IEEEtran}
\bibliography{references}

\end{document}